\PassOptionsToPackage{table}{xcolor}
\documentclass[10pt,twocolumn,letterpaper]{article}

\usepackage[pagenumbers]{wacv} % To force page numbers, e.g. for an arXiv version

\definecolor{wacvblue}{rgb}{0.21,0.49,0.74}
\usepackage[pagebackref,breaklinks,colorlinks,allcolors=wacvblue]{hyperref}
\usepackage{multirow}

\definecolor{ourgray}{gray}{0.9}

\def\wacvPaperID{2495} % *** Enter the WACV Paper ID here
\def\confName{WACV}
\def\confYear{2027}

\title{Predicting Metastatic Risk from Primary Cancer Tissue Architecture via Distance-Aware Spatial Modeling}

\author{
Sandesh Pokhrel\\
Department of Electrical and Computer Engineering\\
Scientific Computing and Imaging Institute\\
University of Utah, Salt Lake City, UT, USA\\
{\tt\small u1604138@umail.utah.edu}
\and
Hamid Manoochehri\\
Department of Electrical and Computer Engineering\\
Scientific Computing and Imaging Institute\\
University of Utah, Salt Lake City, UT, USA
\and
Beatrice S. Knudsen\\
Department of Pathology\\
University of Utah, Salt Lake City, UT, USA
\and
Tolga Tasdizen\\
Department of Electrical and Computer Engineering\\
Scientific Computing and Imaging Institute\\
University of Utah, Salt Lake City, UT, USA
}

\begin{document}
\maketitle
\begin{abstract}
Predicting distant metastasis from the digital H \& E slides of the primary tumor is a critical yet challenging task in computational pathology. Multiple Instance Learning (MIL) approaches can attend to subdomains in whole slide images (WSIs) that harbor features of pre-metastatic cancer regions. However, conventional MIL models largely treat tissue patches as unordered bags, discarding the spatial layout that defines how these regions are arranged and interact across the tissue. We propose that metastatic risk is shaped not only by local patch appearance, but also by the geometric organization of patches in the WSI and the interaction between the tissue compartments. To this end, we introduce Distance-aware Tissue Modeling for Multiple Instance Learning (DTMF-MIL), a spatial MIL framework that reinforces feature embeddings with explicit distance priors. By computing signed distance functions (SDFs) to capture regions with similar features, and representing each patch with radial-basis distance responses and local SDF statistics, DTMF-MIL learns positions of patches with respect to regional interiors and boundaries. The interactions between similar patches are contextualized across local tissue neighborhoods and used to guide slide-level attention while pooling patch feature evidence for metastasis prediction. We evaluate DTMF-MIL for prediction of distant prostate cancer metastasis in an internal prostate needle-biopsy cohort (IPC) from a large hospital system and on public TCGA-COAD and TCGA-KIRC datasets across multiple pathology foundation-model backbones. Across most dataset, backbone, and metric combinations, DTMF-MIL achieves the strongest results consistently. Additional analyses using segmentation-derived epithelial and stromal partitions suggest that coarse tissue-interface geometry contains useful metastatic-risk signal, consistent with the biological role of stromal-epithelial interactions in cancer progression.
\end{abstract}
    
\section{Introduction}
\label{sec:intro}

Cancer metastasis accounts for a large fraction of cancer mortality, making prediction of metastasis from primary tumor tissue an important problem in computational pathology \cite{Metastases}. Predicting it directly from primary histology is valuable from a clinical perspective as biopsy and resection tissue are often available at diagnosis, before distant disease is clinically established, supporting earlier risk stratification and treatment planning without requiring metastatic-site tissue. Unlike tumor detection or cancer subtyping, distant-metastasis labels require downstream clinical staging, follow-up, imaging, or pathology reports, and are therefore not readily available at scale, making the task of metastasis prediction uncommon.

Whole-slide images (WSIs) are too large to process with standard computer vision techniques, so computational pathology models commonly decompose them into patch embeddings and learn from slide-level labels using Multiple Instance Learning (MIL). This formulation has been highly effective because it avoids the need for dense pathologist annotations and allows attention to identify discriminative regions \cite{abmil,clam,dsmil,dtfdmil}. These models are well suited when the label is well explained by a small set of visually diagnostic tiles. Metastatic progression is less likely to be reducible to a small set of diagnostic patches because two patches with similar morphology may carry different meaning depending on whether they lie inside tumor-rich regions, near stromal interfaces, or within heterogeneous tissue neighborhoods. Unlike diagnostic tasks that can often hinge on a localized histologic pattern, metastatic progression is shaped by interactions between malignant cells and the surrounding tissue environment. Tumor nests, stromal context, immune infiltration, epithelial organization, and the geometry of tissue interfaces can influence invasion and dissemination \cite{metastasisandphenotypes,metastatic_colonization,tissue_geometry_morphogenesis,metastatic_state_profiling,CTCmetastasis}. This is especially relevant for prostate cancer, where micro-dissected epithelial and stromal transcriptional profiling has shown that stromal programs vary with tumor grade and metastatic potential \cite{prostate_stromal_epithelial_map}. More broadly, stromal--epithelial interaction and cancer--stroma crosstalk are recognized contributors to tumor progression and metastatic colonization \cite{stromal_epithelial_interaction_review,cancer_stroma_crosstalk_colonization}. Metastasis prediction therefore provides a useful setting for asking whether a model can reason over tissue architecture and tissue-compartment interactions, as well as patch appearance.

Recent MIL methods attempt to recover some of this missing context. Transformer-based models such as TransMIL model correlations among instances, region-aware methods such as RRT-MIL re-embeds patch features into more contextual representations, and graph-based methods such as WiKG encode relations between patch nodes \cite{transmil,rrtmil,patchgcn,wikg}. These works motivate the importance of context, but they often model patch-to-patch similarity, sequence positioning or graph connectivity. Such mechanisms do not explicitly represent where a patch lies relative to phenotypic interiors and boundaries where signs of invasion can live. For metastatic risk prediction, this distinction matters because the relevant signal may lie in how phenotypic regions interact rather than simply the presence of an isolated high-attention patch. 

In this paper, we introduce Distance-aware Tissue Modeling for Multiple Instance Learning (DTMF-MIL), a spatial MIL framework for metastasis prediction from primary tissue histology. DTMF-MIL partitions patch embeddings into tissue regions and computes signed distance functions (SDFs) relative to each region. Five radial-basis distance responses and local SDF statistics encode whether a patch lies within, near, or far from a tissue partition boundary. These spatial descriptors interact with patch features before attention pooling, allowing the model to aggregate feature evidence conditioned on tissue geometry.

We also examine how the granularity and origin of tissue partitions affect this spatial signal. Alongside K-means partitions at different values of $K$, we evaluate two-region SDFs obtained from a prostate-tissue segmentation model \cite{segmentationmodel} that estimates epithelial (tumor) and stromal regions. This analysis provides an interpretable reference for testing whether a biologically recognizable tissue compartment is informative for metastatic risk prediction, while the multi-region K-means analysis tests whether abstract tissue phenotypes capture additional spatial structure. Our results suggest that even coarse tissue-interface geometry contains useful signal to predict metastatic potential.

To our knowledge, this is the first MIL framework to utilize phenotype region distance fields as explicit spatial priors for metastatic risk prediction. Overall, our contributions in this work can be summarized as:
\begin{enumerate}
    \item We propose DTMF-MIL, a novel distance-aware spatial MIL framework that uses SDF-derived tissue-geometry features to guide metastasis prediction from primary tissue histology.
    \item We evaluate DTMF-MIL across an internal prostate cancer needle-biopsy cohort (IPC) and TCGA-COAD/TCGA-KIRC cohorts, showing consistently top-tier performance across multiple foundation model backbones (UNI/UNI2H/Phikon-v2).
    \item We analyze K-means partition granularity and model-derived epithelial/stromal SDFs, showing that both coarse two-region and richer multi-region partitions can provide useful spatial inputs.
\end{enumerate}

\section{Related Work}
\label{sec:related_works}

\textbf{Multiple instance learning in pathology.}
MIL is the standard weakly supervised formulation for WSI classification because it maps a bag of image patch embeddings to a slide-level label without requiring patch-level annotations \cite{abmil,clam}. Attention-based MIL learns instance weights that can highlight discriminative patches, while CLAM adds clustering-based supervision to encourage class-specific instance separation \cite{abmil,clam}. DSMIL extends this idea with two streams: an instance classifier selects a critical patch, and a bag classifier aggregates other patches according to their similarity to that critical instance \cite{dsmil}. DTFD-MIL instead divides the bag into pseudo-bags and distills their features through a second aggregation stage, improving feature utilization in large WSIs \cite{dtfdmil}. These methods provide strong baselines but they critically miss the relationship between instances and how microenvironment in neighboring patches can influence each other.

\textbf{Contextual, transformer, and graph MIL.}
A second family of MIL methods models relationships between instances. TransMIL uses Transformer-style correlated instance modeling with positional encoding, while RRT-MIL re-embeds features to improve contextual aggregation \cite{transmil,rrtmil}. ILRA \cite{ilra} uses iterative low rank attention to encode cross-instance correlation. Graph-based methods such as Patch-GCN and WiKG represent patches as nodes and use graph connectivity or knowledge-aware attention to propagate information across related instances \cite{patchgcn,wikg}. These approaches are more relational than independent attention pooling, but their relations are still typically learned from feature affinity, graph construction, or positional ordering. Other MIL methods make the role of coordinates more explicit by introducing spatial context into attention \cite{PSAMIL} and using approximate convolutional conditional random field to generate spatial encoding \cite{bayesmil}. 

These designs show that spatial context can improve WSI modeling, but they commonly encode spatial relationships as patch-to-patch proximity, sequence position, or smoothed neighborhood interaction. For metastatic risk however, our assumption is different: tissue geometry is informative because a patch's meaning can depend on its distance to tissue-region interiors and interfaces. A patch that is otherwise visually uninformative may be of importance because it lies near an invasive interface, between two tissue compartments, or within a spatial neighborhood that changes the meaning of its morphology. This motivates spatial representations that describe tissue organization directly rather than relying only on spatial similarities alone. DTMF-MIL therefore represents patch location relative to data-driven tissue-region boundaries through SDFs, making tissue-interface geometry an explicit input to attention.

\textbf{Spatial structure and metastatic risk.}
Tumor progression and metastasis are influenced by the organization of cancer cells and the surrounding microenvironment \cite{metastasisandphenotypes,metastatic_colonization,tissue_geometry_morphogenesis,cell_geometry}. In prostate cancer, Tyekucheva et al. generated stromal and epithelial transcriptional maps and reported that stromal programs differed across disease states and were associated with aggressive and metastatic potential \cite{prostate_stromal_epithelial_map}. This supports a central motivation of our work: metastatic-risk information may be present in the context surrounding tumor epithelium, not only in the most critical epithelial tiles. Broader cancer biology also emphasizes stromal--epithelial interaction and cancer--stroma crosstalk as drivers of progression and metastatic colonization \cite{stromal_epithelial_interaction_review,cancer_stroma_crosstalk_colonization}.  Prior computational pathology work has similarly shown that tissue phenotype composition and neighborhood statistics can be predictive of distant metastasis, supporting the view that risk is not determined by tumor cells alone \cite{metastasisandphenotypes}. In histology, this motivates models that can distinguish not only what tissue phenotypes are present, but where they occur relative to one another. SDFs provide a continuous way to encode this relationship: each patch is described by its distance to multiple tissue regions and their boundaries rather than by a discrete cluster identity alone.  

This representation gives the MIL model a geometry-aware coordinate system over phenotypic tissue structures that can be derived through pathology foundation models using K-means or through recent unsupervised histopathology segmentation methods that can segment biologically interpretable tissue partitions. We use the epithelial/stromal outputs of segmentation model (Cai et al. \cite{segmentationmodel}) as an annotation-like reference for IPC dataset analysis, because this partition aligns with the stromal--epithelial biology. We do not treat these labels as pathologist ground truth; they serve as a controlled test of whether a recognizable tissue interface can support the same distance-field modeling principle used with K-means partitions.

\textbf{Pathology foundation models.}
Recent pathology foundation models provide strong patch embeddings for downstream MIL tasks. We evaluate DTMF-MIL with multiple backbones, including UNI, UNI2H \cite{uni2h} and Phikonv2 \cite{phikonv2}, to test whether explicit spatial encoding remains useful when patch features are already strong. This is important because spatial features should add information beyond what the foundation model already encodes.

\section{Methodology}
\label{sec:method}

\begin{figure*}[t]
    \centering
    \includegraphics[width=\linewidth]{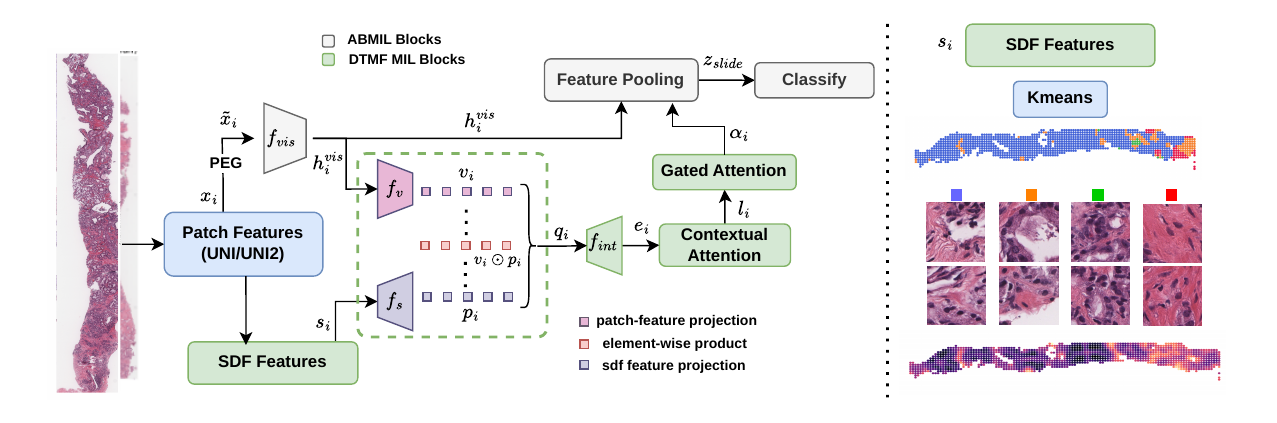}
    \caption{\textbf{DTMF-MIL overview.} We illustrate the workflow of DTMF-MIL in the left panel. Foundation-model patch embeddings are used both as visual inputs and to obtain fold-specific K-means tissue partitions. The gray blocks illustrate conventional ABMIL pipeline which only utilizes the patch feature embeddings. We reinforce these patch feature embeddings with distance features derived from data-derived phenotypic clustering. Projected visual and spatial features, together with their element-wise product, are contextualized by the attention module. The resulting attention weights pool the visual patch features for slide-level classification. The right panel shows the extraction of SDF features from the UNI/UNI2H patch features and samples of patches from their respective clusters. An overall magnitude map of the spatial features is also presented. More details on the specific SDF-features are illustrated in figure \ref{fig:sdf_features}.}
    \label{fig:method}
\end{figure*}

In MIL, a whole-slide image (WSI) is represented as a variable-length bag of patch embeddings with associated grid coordinates,
\begin{equation}
\begin{aligned}
    \mathcal{B} &= \{(x_i,\mathbf{p}_i)\}_{i=1}^{N}, \\x_i &\in \mathbb{R}^{D}, \\
    \mathbf{p}_i &= (u_i,v_i), \\
    Y &\in \{0,1\}.
\end{aligned}
\label{eq:wsi-bag}
\end{equation}
Here, $x_i$ is a foundation-model visual embedding, $\mathbf{p}_i$ is the patch-grid coordinate, and $Y$ denotes the metastasis label. Conventional MIL learns a slide-level prediction from the visual bag alone, $\hat{Y}=\rho\!\left(\phi\left(\{x_i\}_{i=1}^{N}\right)\right)$
% \begin{equation}
%     \hat{Y}=\rho\!\left(\phi\left(\{x_i\}_{i=1}^{N}\right)\right),
%     \label{eq:mil-formulation}
% \end{equation}
where $\phi$ aggregates instances and $\rho$ maps the slide representation to the output label. DTMF-MIL keeps this weakly supervised setting, but augments each instance with a spatial descriptor $s_i$ derived from the arrangement of tissue regions. The spatial descriptor provides geometric context that helps the model determine where evidence of metastatic potential could exist.

\subsection{Tissue Regions and Signed Distance Fields}

We first obtain a slide-level partition of patch embeddings into $K$ data-derived tissue phenotypic regions. To avoid leakage from held-out slides, this partitioning is learned within each cross-validation fold: K-means is fitted only on patch features from the training patients for that fold, and the centroids are then used to assign patches from validation and test patients. Thus, a three-fold experiment uses three fold-specific K-means models rather than a single K-means model fitted on the full cohort. The K-means is fitted on the training set so that patches across different WSIs can be mapped into consistent phenotypic regions. Each patch is assigned to a region as, $k_i \in \{1,\ldots,K\}.$
For each region $k$, we define a binary mask $M_k$ on the patch grid, $M_k(u_i,v_i)=\mathbf{1}[k_i=k]$.
This mask is used to construct a signed distance map using
\begin{equation}
    \psi_k(u,v)
    =\operatorname{EDT}(1-M_k)(u,v)
    -\operatorname{EDT}(M_k)(u,v),
    \label{eq:sdf}
\end{equation}
where $\operatorname{EDT}$ is the Euclidean distance transform used to calculate the \textit{signed distance function} (SDF). SDF values are near zero at tissue-region interfaces and changes sign across the region boundary i.e $\psi_k(u,v)>0$ outside the tissue region and $\psi_k(u,v)<0$ inside it. Thus, the SDF collection gives every patch a continuous description of its location relative to all tissue regions, rather than a discrete region identity alone.  This is important because a cluster label only says what region a patch belongs to, whereas the SDF also distinguishes region interiors, boundary neighborhoods, and exterior tissue. The main experiments use $K=5$ regions. Details of different phenotypic regions reviewed by a pathologist are discussed in the supplementary section \ref{sec:vlm_validation}.

\textbf{Segmentation based regions.} Because stromal--epithelial organization has biological relevance for prostate cancer progression and metastatic potential \cite{prostate_stromal_epithelial_map,stromal_epithelial_interaction_review}, the IPC dataset analysis also considers two-region partitions: epithelial/stromal regions estimated by an unsupervised histopathology semantic segmentation model, Cai et al.~\cite{segmentationmodel}. The model uses overlapping-patch consistency and inter-layer self-attention fusion to produce context-aware semantic tissue regions. We use its epithelial/stromal output only to construct two-region SDF masks for analysis; the segmentation labels are not used as supervision for metastasis prediction. Since the model produces pixel-level segmentation, each patch is assigned to epithelium or stroma by majority pixel label within the patch area. This serves as a alternative of K-means with 2 clusters for biologically aligned tissue phenotypes. We go into the details of the segmentation model in the supplementary section \ref{sec:segmentation_model}.

\subsection{SDF-Derived Spatial Features}

For patch $i$ and region $k$, we sample the SDF at the patch coordinate, $d_i^{(k)}=\psi_k(u_i,v_i)$.
% \begin{equation}
%     d_i^{(k)}=\psi_k(u_i,v_i).
%     \label{eq:patch-distance}
% \end{equation}
Raw signed distances are converted into five radial-basis function (RBF) responses,
\begin{equation}
\begin{aligned}
    r_{i,j}^{(k)}
    &=\exp\!\left(-\lambda_j\left(d_i^{(k)}\right)^2\right), & \lambda_j \in \Lambda, \ |\Lambda| = 5
\end{aligned}
\label{eq:rbf}
\end{equation}
where $\Lambda$ is a set of hyperparameters selected to model multi-scale responses.
% with
% \begin{equation}
% \begin{aligned}
%     \Lambda=\{&10^{-7},\;3.59\!\times\!10^{-7},\;1.29\!\times\!10^{-6}, \\
%               &4.64\!\times\!10^{-6},\;1.67\!\times\!10^{-5}\}.
% \end{aligned}
% \label{eq:rbf-lambdas}
% \end{equation}
Using multiple radial bases is a deliberate bias toward multi-scale tissue geometry rather than a single hand-tuned boundary width. Smaller $\lambda_j$ values vary slowly over broader tissue extent, whereas larger values change more sharply near cluster boundaries.

Since, distance alone does not capture the nature of cluster region boundaries. We complement the RBF responses with horizontal and vertical SDF gradients,
\begin{equation}
\begin{aligned}
    g_{x,i}^{(k)} &= \nabla_x\psi_k(u_i,v_i), & g_{y,i}^{(k)} = \nabla_y\psi_k(u_i,v_i),
\end{aligned}
\label{eq:sdf-gradients}
\end{equation}
and with local SDF statistics (SDF mean and standard deviation) in a $3\times3$ neighborhood $\mathcal{N}_3(i)$,
\begin{equation}
\begin{aligned}
    \mu_i^{(k)}
    &=\frac{1}{|\mathcal{N}_3(i)|}
      \sum_{q\in\mathcal{N}_3(i)}\psi_k(q), \\
    \sigma_i^{(k)}
    &=\operatorname{Std}_{q\in\mathcal{N}_3(i)}[\psi_k(q)].
\end{aligned}
\label{eq:local-sdf}
\end{equation}

\begin{figure*}[t]
    \centering
    \includegraphics[width=\linewidth]{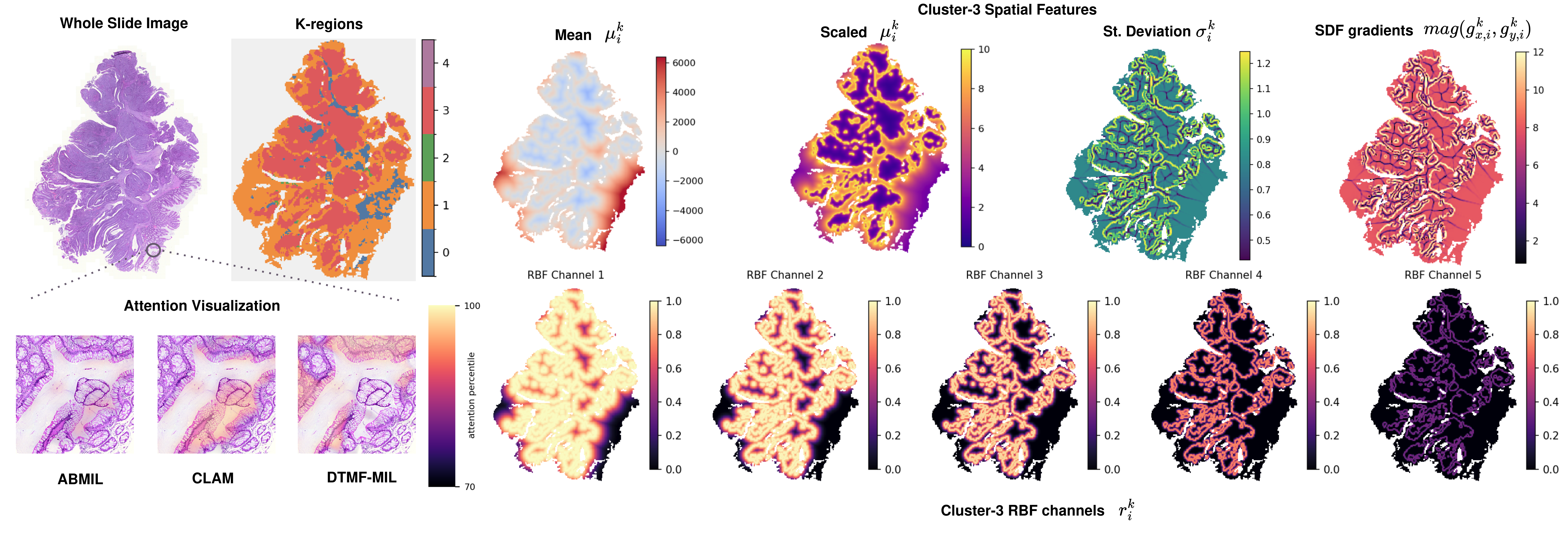}
    \caption{Example TCGA-COAD WSI with \(K=5\) tissue partitions and SDF-derived features for cluster 3. From left to right, we show the WSI, cluster assignments, raw local mean, scaled local mean, local standard deviation, gradient magnitude, and five RBF channels. The bottom-left panels compare ABMIL, CLAM, and DTMF-MIL-Local attention on the same tissue region. Visualizations of attention for other slides are provided in the supplementary section \ref{sec:attn_viz}.}
    \label{fig:sdf_features}
\end{figure*}

Following this, the per-region distance descriptor is, 
\begin{equation}
\begin{aligned}
    s_i^{(k)}= [&r_{i,1}^{(k)},\ldots,r_{i,5}^{(k)},
    g_{x,i}^{(k)},g_{y,i}^{(k)}, \\
    &\mu_i^{(k)},\sigma_i^{(k)}],
\end{aligned}
\label{eq:per-region-feature}
\end{equation}
and the full spatial descriptor concatenates all regions as,
\begin{equation}
\begin{aligned}
    s_i &= s_i^{(1)}\Vert\cdots\Vert s_i^{(K)}, & s_i \in \mathbb{R}^{9K}.
\end{aligned}
\label{eq:full-spatial-feature}
\end{equation}
With $K=5$, DTMF-MIL uses a 45-dimensional spatial descriptor per patch. The SDF mean values are scaled to a range of 0-10 as appropriate input values to the network instead of the large magnitude ($10^3$) values that can make the network unstable; details of the scaling function are attached in the supplementary section \ref{sec:scaling}.
The resulting spatial descriptor is compact yet still separates three forms of geometry: distance to tissue regions through RBFs, local interface direction through gradients, and neighborhood heterogeneity through local statistics. This rich representation captures a comprehensive view of where the patch resides relative to every identified tissue phenotype, conditioned on both distance and local structural variation. Figure \ref{fig:sdf_features} provides a visual overview of the resulting spatial features.

\subsection{Visual-Spatial Interaction and Contextual Attention}

The patch features from the foundation models are passed through a learnable Position Encoding Generator (PEG) module following RRTMIL, Transmil \cite{rrtmil,transmil} as they capture learnable spatial significance per patch. PEG provides generic coordinate-based positional encoding and is distinct from our phenotype-derived spatial representation. This allows the contribution of the proposed distance-aware features to be evaluated independently while retaining the same positional encoder. The encoded patch features are then mapped through a two-layer feature extractor ($f_{\mathrm{vis}}$),
\begin{equation}
\begin{aligned}
    \tilde{x}_i &= \operatorname{PEG}(x_i), & h_i^{\mathrm{vis}} &= f_{\mathrm{vis}}(\tilde{x}_i),  &h_i^{\mathrm{vis}} \in \mathbb{R}^{512}
\end{aligned}
\label{eq:visual-embedding}
\end{equation}
The patch feature and spatial streams are projected separately into a 64-dimensional interaction space as $v_i = f_v(h_i^{\mathrm{vis}}), v_i \in \mathbb{R}^{64}$ and $p_i = f_s(s_i), p_i \in \mathbb{R}^{64}.$
% \begin{equation}
% \begin{aligned}
%     v_i &= f_v(h_i^{\mathrm{vis}}), & v_i &\in \mathbb{R}^{64}, \\
%     p_i &= f_s(s_i),               & p_i &\in \mathbb{R}^{64}.
% \end{aligned}
% \label{eq:projections}
% \end{equation}
Projecting both modalities to the same compact dimension allows their channels to be compared without allowing the much higher-dimensional patch feature embedding to dominate the spatial descriptor. We use this bottleneck because the spatial input is intended to condition patch feature interpretation, not to add a large parallel feature space that can overfit to partition artifacts. We form the interaction by concatenating this patch feature context (visual context), spatial context, and their element-wise product.
\begin{equation}
\begin{aligned}
    q_i &= [v_i\;\Vert\;p_i\;\Vert\;(v_i\odot p_i)], & q_i &\in \mathbb{R}^{192}, \\
    e_i &= f_{\mathrm{int}}(q_i),  & e_i &\in \mathbb{R}^{512}.
\end{aligned}
\label{eq:vxp}
\end{equation}
Here, $\odot$ denotes an element-wise product. The product term exposes alignment between visual and spatial channels. This makes the interaction explicitly conditional: visually similar patch could contribute differently depending on whether it appears inside a region, near an interface, or in a heterogeneous local neighborhood. The interaction projector ($f_{int}$) uses a 128-dimensional hidden layer before returning to the 512-dimensional token space. Figure \ref{fig:method} illustrates these components for clarity.

\textbf{Contextual Attention}. The DTMF-MIL-Local model contextualizes these interaction tokens $e_i$ with four layers of four-head self-attention within a local 16x16 patch windows. The windowed design reflects the expectation that many tissue-interface effects are locally organized, while avoiding the computational cost and instability of full slide-level attention over thousands of patches. This allows tissue geometry to influence a patch through its nearby context while retaining tractable slide-level computation. The shifted-window (DTMF-MIL-Swin) broadens communication across adjacent windows by alternating the window offset between layers. DTMF-MIL-Nys. approximates global self-attention on all the patches with Nystr{\"o}m attention, testing whether slide level contextual mixing changes the utility of the same spatial representation.

The contextualization module preserves the patch-token structure: each interaction token $e_i$ is updated into a contextualized interaction token $c_i$. We then use these contextualized tokens to compute patch-importance weights. Specifically, a gated-attention scoring network $a(\cdot)$ maps each $c_i$ to a scalar logit,
\begin{equation}
\begin{aligned}
    c_i &= \operatorname{Context}(e_i), \\
    \ell_i &= a(c_i), \\
    \alpha_i &= \frac{\exp(\ell_i)}{\sum_{j=1}^{N}\exp(\ell_j)}, \\
    % \sum_{i=1}^{N}\alpha_i &= 1.
\end{aligned}
\label{eq:attention-weights}
\end{equation}
The final slide representation is formed by applying these spatially informed weights to the patch features,
\begin{equation}
\begin{aligned}
    z_{\mathrm{slide}} &= \sum_{i=1}^{N}\alpha_i h_i^{\mathrm{vis}}, &
    \hat{Y} = \rho(z_{\mathrm{slide}}).
\end{aligned}
\label{eq:pooling}
\end{equation}

Spatial features therefore guide \emph{where} the model looks for visual signs of invasion, while the pooled slide representation comes from a foundation model derived patch features. This separation preserves the patch feature evidence used for prediction while making its aggregation sensitive to tissue geometry.

\section{Experiments and Results}
\label{sec:experiments}

\subsection{Experimental Setup}
\begin{table*}[t]
    \centering
    \footnotesize
    \setlength{\tabcolsep}{4pt}
    \renewcommand{\arraystretch}{1.08}
    \caption{\textbf{IPC metastatic-risk prediction with UNI backbone.} Bold indicates the best result in each metric and underline indicates the second best result.}
    \label{tab:va_uni_results}
    \begin{tabular}{lccccc}
        \toprule
        Method & BAcc$\uparrow$ & AUC$\uparrow$ & F1$\uparrow$ & Accuracy$\uparrow$ & $\kappa{\times}100$ \\
        \midrule
        \multicolumn{6}{l}{\textit{Spatial methods (Ours)}} \\
        \rowcolor{ourgray}
        DTMF-MIL-Local & $\mathbf{82.97 \pm 03.03}$ & $\underline{90.46 \pm 01.03}$ & $\mathbf{82.59 \pm 03.46}$ & $\mathbf{82.69 \pm 03.19}$ & $\mathbf{65.50 \pm 06.18}$ \\
        \rowcolor{ourgray}
        DTMF-MIL-Swin & \underline{$81.38 \pm 03.56$} & $90.00 \pm 01.43$ & $\underline{81.21 \pm 03.56}$ & $\underline{81.08 \pm 03.69}$ & $\underline{62.30 \pm 07.15}$ \\
        \rowcolor{ourgray}
        DTMF-MIL-Nystrom & $80.61 \pm 03.61$ & $90.14 \pm 01.48$ & $80.77 \pm 04.57$ & $80.57 \pm 03.58$ & $61.07 \pm 07.14$ \\
        % \addlinespace
        \multicolumn{6}{l}{\textit{Non-spatial baselines}} \\
        \hline
        CLAM & $79.23 \pm 04.66$ & $\mathbf{90.57 \pm 01.21}$ & $79.53 \pm 05.02$ & $79.22 \pm 04.33$ & $58.29 \pm 09.01$ \\
        WiKG & $78.62 \pm 03.61$ & $89.52 \pm 01.85$ & $77.65 \pm 04.89$ & $78.52 \pm 03.56$ & $57.01 \pm 07.23$ \\
        ABMIL & ${79.67 \pm 04.13}$ & $89.81 \pm 01.99$ & $80.26 \pm 04.83$ & $79.40 \pm 04.26$ & $58.95 \pm 08.28$ \\
        DSMIL & $77.99 \pm 08.33$ & $89.20 \pm 01.57$ & $78.97 \pm 05.91$ & $77.60 \pm 08.73$ & $55.58 \pm 16.72$ \\
        RRT-MIL & $78.12 \pm 04.30$ & $89.22 \pm 01.72$ & $76.30 \pm 07.40$ & $77.75 \pm 04.46$ & $55.74 \pm 08.59$ \\
        ILRA & $78.35 \pm 04.00$ & $88.08 \pm 02.83$ & $78.77 \pm 04.33$ & $78.47 \pm 04.01$ & $56.73 \pm 08.00$ \\
        TransMIL & $77.07 \pm 04.32$ & $88.69 \pm 02.58$ & $75.91 \pm 10.16$ & $76.91 \pm 04.78$ & $53.94 \pm 08.90$ \\
        DFTD-MIL & $72.79 \pm 08.11$ & $79.01 \pm 07.15$ & $70.42 \pm 12.75$ & $72.45 \pm 08.41$ & $45.26 \pm 16.20$ \\
        \bottomrule
    \end{tabular}
\end{table*}

\noindent\textbf{Cohorts and prediction task.} To test whether distance-aware tissue geometry is useful both in the target clinical setting and beyond it, we evaluate DTMF-MIL on two complementary cohorts: an internal prostate needle-biopsy cohort (IPC) from a large hospital system for primary tissue distant metastasis prediction, and public TCGA  colon adenocarcinoma (TCGA-COAD) and TCGA kidney renal clear cell carcinoma (TCGA-KIRC) \cite{tcga} cohorts that provide more imbalanced external settings across distinct tissue types. The IPC cohort contains 502 WSIs and serves as the primary evaluation setting because it has a comparatively balanced outcome distribution with 241 M0 and 261 M1 slides. For TCGA, distant-metastasis labels are derived from the pathological TNM M category following the TCGA pathology-report staging resource of Kefeli and Tatonetti \cite{tnmstage}. We binarize M status as M0 versus M1, where M1 denotes distant metastasis. After filtering for available labels, TCGA-COAD contains 370 labelled slides (309 M0, 61 M1), and TCGA-KIRC contains 476 labelled slides (403 M0, 73 M1). These cohorts allow us to assess whether the spatial representation remains useful beyond the IPC cohort and across more class-imbalanced disease settings. We report results with UNI,UNI2H and Phikon-v2 visual foundation-model features. We will release the public dataset cohort details for reproducibility. Technical implementation details are attached in the supplementary section \ref{sec:implementation_details}.

\noindent\textbf{Evaluation protocol and metrics.}
All main comparisons use five random seeds and patient-level three-fold cross-validation; values are the mean $\pm$ standard deviation over five seed-level cross-validation summaries. The same patient-level splits are used across methods within each cohort/backbone setting. On the comparatively balanced IPC cohort, we report AUC, F1-score, accuracy, and BAcc. We also report Cohen's Kappa which measures observed agreement beyond the agreement expected by chance under the empirical class prevalences: $\kappa=0$ indicates chance-level agreement and $\kappa=1$ perfect agreement. In our setting, a $\kappa$ near 1 means that the model almost perfectly separates metastatic from non-metastatic slides, with agreement far beyond chance. For compact display, all table entries are multiplied by 100; $\kappa$ is unitless and should be interpreted after dividing by 100. For the imbalanced TCGA cohorts, we report AUC and BAcc as these metrics are less dominated by majority class prevalence than raw accuracy.

\noindent\textbf{Baselines.}
We compare against ABMIL, CLAM, DSMIL, RRT-MIL, ILRA, TransMIL, WiKG, and DFTD-MIL \cite{abmil,clam,dsmil,rrtmil,ilra,transmil,wikg,dtfdmil}. \textit{DTMF-MIL-Local} is the primary 16x16 window local-context attention model. \textit{DTMF-MIL-Swin} and \textit{DTMF-MIL-Nystrom} use the same SDFs and feature pooling mechanism while probing shifted-window and approximate global contextual mixing, respectively.

\begin{table*}
    \centering
    \footnotesize
    \setlength{\tabcolsep}{4pt}
    \renewcommand{\arraystretch}{1.08}
    \caption{\textbf{IPC metastatic-risk prediction with UNI2H backbone.} Bold indicates the best result in each metric and underline indicates the second best result.}
    \label{tab:va_uni2h_results}
    \begin{tabular}{lccccc}
        \toprule
        Method & BAcc$\uparrow$ & AUC$\uparrow$ & F1$\uparrow$ & Accuracy$\uparrow$ & $\kappa{\times}100$ \\
        \midrule
        \multicolumn{6}{l}{\textit{Spatial methods (Ours)}} \\
        \rowcolor{ourgray}
        DTMF-MIL-Local & $\mathbf{83.07 \pm 02.24}$ & $\mathbf{91.58 \pm 01.05}$ & $\underline{82.47 \pm 02.72}$ & $\mathbf{82.86 \pm 02.28}$ & $\mathbf{65.77 \pm 04.50}$ \\
        \rowcolor{ourgray}
        DTMF-MIL-Swin & $\underline{82.85 \pm 02.75}$ & $\underline{91.40 \pm 01.18}$ & $\mathbf{82.69 \pm 03.31}$ & $\underline{82.77 \pm 02.71}$ & $\underline{65.49 \pm 05.42}$ \\
        \rowcolor{ourgray}
        DTMF-MIL-Nystrom & $81.62 \pm 06.78$ & $91.39 \pm 01.10$ & $82.25 \pm 03.69$ & $81.54 \pm 06.32$ & $62.92 \pm 13.37$ \\
        % \addlinespace
        
        \multicolumn{6}{l}{\textit{Non-spatial baselines}} \\
        \hline
        CLAM & $82.09 \pm 02.96$ & $90.99 \pm 01.84$ & $82.14 \pm 04.20$ & $82.06 \pm 03.05$ & $64.06 \pm 05.93$ \\
        WiKG & $81.15 \pm 02.63$ & $90.73 \pm 01.68$ & $81.25 \pm 02.99$ & $80.95 \pm 02.83$ & $61.96 \pm 05.44$ \\
        ABMIL & $81.44 \pm 03.47$ & $90.61 \pm 01.44$ & $81.63 \pm 03.30$ & $81.25 \pm 03.53$ & $62.55 \pm 06.89$ \\
        DSMIL & $79.28 \pm 04.28$ & $88.65 \pm 03.24$ & $77.49 \pm 06.85$ & $78.94 \pm 04.52$ & $58.10 \pm 08.75$ \\
        RRT-MIL & $78.40 \pm 03.96$ & $89.58 \pm 02.15$ & $76.98 \pm 06.72$ & $78.18 \pm 04.12$ & $56.48 \pm 07.90$ \\
        ILRA & $74.78 \pm 10.21$ & $88.68 \pm 02.05$ & $74.06 \pm 15.86$ & $75.29 \pm 09.83$ & $49.80 \pm 20.26$ \\
        TransMIL & $76.46 \pm 06.29$ & $88.26 \pm 02.16$ & $76.94 \pm 08.23$ & $76.52 \pm 05.74$ & $52.73 \pm 12.39$ \\
        DFTD-MIL & $76.93 \pm 04.51$ & $83.49 \pm 03.87$ & $76.23 \pm 07.95$ & $76.62 \pm 04.75$ & $53.46 \pm 09.06$ \\
        \bottomrule
    \end{tabular}
\end{table*}

\subsection{Comparative Results}
\noindent\textbf{IPC dataset results.}
Tables~\ref{tab:va_uni_results} and~\ref{tab:va_uni2h_results} provide the complete IPC  comparison. With UNI, DTMF-MIL-Local attains the best F1-score, accuracy, BAcc, and Cohen's kappa; CLAM has a marginally higher AUC. With UNI2H, DTMF-MIL-Local achieves the best AUC, accuracy, BAcc, and Cohen's kappa, while DTMF-MIL-Swin has the best F1-score. We also present results on Phikon-V2 backbone in supplementary section \ref{sec:ipc_phikon-v2}.

\begin{table}[t]
    \centering
    \scriptsize
    \setlength{\tabcolsep}{2.0pt}
    \renewcommand{\arraystretch}{1.04}
    \caption{\textbf{TCGA-COAD comparison} across UNI and UNI2H.}
    \label{tab:tcga_coad_results}

    \resizebox{\columnwidth}{!}{%
    \begin{tabular}{@{}llcc@{}}
        \toprule
        Backbone & Method & AUC$\uparrow$ & BAcc$\uparrow$ \\
        \midrule

        \multirow{11}{*}{\textbf{UNI}}
        & \cellcolor{ourgray} DTMF-MIL-Nys.
        & \cellcolor{ourgray} $\mathbf{74.03{\pm}04.38}$
        & \cellcolor{ourgray} $\mathbf{64.71{\pm}04.59}$ \\
        & \cellcolor{ourgray} DTMF-MIL-Local
        & \cellcolor{ourgray} $73.29{\pm}04.26$
        & \cellcolor{ourgray} $63.88{\pm}05.99$ \\
        & \cellcolor{ourgray} DTMF-MIL-Swin
        & \cellcolor{ourgray} $\underline{73.91{\pm}04.67}$
        & \cellcolor{ourgray} $\underline{64.53{\pm}05.60}$ \\
        & CLAM     & $72.70{\pm}04.97$ & $62.50{\pm}03.64$ \\
        & WiKG     & $64.57{\pm}07.08$ & $59.09{\pm}06.29$ \\
        & ABMIL    & $70.09{\pm}03.12$ & $60.56{\pm}04.95$ \\
        & DSMIL    & $62.83{\pm}04.44$ & $55.50{\pm}04.78$ \\
        & RRT-MIL  & $70.85{\pm}03.08$ & $62.06{\pm}04.26$ \\
        & ILRA     & $65.56{\pm}07.35$ & $58.49{\pm}05.30$ \\
        & TransMIL & $59.03{\pm}06.52$ & $53.97{\pm}04.60$ \\
        & DFTD-MIL & $49.95{\pm}06.84$ & $50.04{\pm}03.95$ \\

        \specialrule{\lightrulewidth}{0.8pt}{0.6pt}
        \specialrule{\lightrulewidth}{0pt}{0.8pt}

        \multirow{11}{*}{\textbf{UNI2H}}
        & \cellcolor{ourgray} DTMF-MIL-Local
        & \cellcolor{ourgray} $\mathbf{72.56{\pm}04.60}$
        & \cellcolor{ourgray} $\mathbf{67.37{\pm}04.54}$ \\
        & \cellcolor{ourgray} DTMF-MIL-Swin
        & \cellcolor{ourgray} $\underline{72.41{\pm}04.51}$
        & \cellcolor{ourgray} $\underline{66.30{\pm}04.58}$ \\
        & \cellcolor{ourgray} DTMF-MIL-Nys.
        & \cellcolor{ourgray} $71.20{\pm}04.69$
        & \cellcolor{ourgray} $64.85{\pm}05.75$ \\
        & CLAM     & $69.84{\pm}03.82$ & $61.40{\pm}04.21$ \\
        & WiKG     & $67.26{\pm}05.98$ & $59.85{\pm}04.09$ \\
        & ABMIL    & $71.19{\pm}05.15$ & $62.35{\pm}05.17$ \\
        & DSMIL    & $63.44{\pm}09.47$ & $56.85{\pm}06.46$ \\
        & RRT-MIL  & $68.21{\pm}06.64$ & $61.65{\pm}03.75$ \\
        & ILRA     & $65.79{\pm}05.49$ & $59.69{\pm}03.82$ \\
        & TransMIL & $54.90{\pm}07.37$ & $53.53{\pm}05.61$ \\
        & DFTD-MIL & $56.66{\pm}07.62$ & $53.11{\pm}04.83$ \\
        \bottomrule
    \end{tabular}%
    }
\end{table}

\begin{table}[t]
    \centering
    \scriptsize
    \setlength{\tabcolsep}{2.0pt}
    \renewcommand{\arraystretch}{1.04}
    \caption{\textbf{TCGA-KIRC comparison} across UNI and UNI2H.}
    \label{tab:tcga_kirc_results}

    \resizebox{\columnwidth}{!}{%
    \begin{tabular}{@{}llcc@{}}
        \toprule
        Backbone & Method & AUC$\uparrow$ & BAcc$\uparrow$ \\
        \midrule

        \multirow{11}{*}{\textbf{UNI}}
        & \cellcolor{ourgray} DTMF-MIL-Swin
        & \cellcolor{ourgray} $\mathbf{83.31{\pm}02.41}$
        & \cellcolor{ourgray} $\mathbf{70.99{\pm}06.23}$ \\
        & \cellcolor{ourgray} DTMF-MIL-Local
        & \cellcolor{ourgray} $82.74{\pm}02.47$
        & \cellcolor{ourgray} $70.04{\pm}06.22$ \\
        & \cellcolor{ourgray} DTMF-MIL-Nys.
        & \cellcolor{ourgray} $81.78{\pm}03.01$
        & \cellcolor{ourgray} $68.06{\pm}05.39$ \\
        & CLAM     & $82.65{\pm}02.70$ & $69.45{\pm}05.59$ \\
        & WiKG     & $\underline{82.75{\pm}02.31}$ & $69.87{\pm}05.95$ \\
        & ABMIL    & $80.25{\pm}03.80$ & $68.09{\pm}05.40$ \\
        & DSMIL    & $80.53{\pm}03.57$ & $66.43{\pm}07.82$ \\
        & RRT-MIL  & $81.68{\pm}03.55$ & $\underline{70.07{\pm}04.88}$ \\
        & ILRA     & $79.77{\pm}03.13$ & $67.99{\pm}06.20$ \\
        & TransMIL & $78.84{\pm}03.62$ & $66.61{\pm}06.29$ \\
        & DFTD-MIL & $68.99{\pm}06.84$ & $62.96{\pm}07.29$ \\

        \specialrule{\lightrulewidth}{0.8pt}{0.6pt}
        \specialrule{\lightrulewidth}{0pt}{0.8pt}

        \multirow{11}{*}{\textbf{UNI2H}}
        & \cellcolor{ourgray} DTMF-MIL-Local
        & \cellcolor{ourgray} $\mathbf{82.64{\pm}02.54}$
        & \cellcolor{ourgray} $68.44{\pm}06.52$ \\
        & \cellcolor{ourgray} DTMF-MIL-Swin
        & \cellcolor{ourgray} $\underline{82.45{\pm}02.54}$
        & \cellcolor{ourgray} $68.89{\pm}06.47$ \\
        & \cellcolor{ourgray} DTMF-MIL-Nys.
        & \cellcolor{ourgray} $81.90{\pm}02.54$
        & \cellcolor{ourgray} $67.29{\pm}06.81$ \\
        & CLAM     & $79.62{\pm}02.88$ & $\underline{69.21{\pm}05.50}$ \\
        & WiKG     & $82.40{\pm}03.73$ & $\mathbf{70.37{\pm}07.89}$ \\
        & ABMIL    & $79.84{\pm}02.77$ & $67.37{\pm}05.68$ \\
        & DSMIL    & $79.87{\pm}03.07$ & $68.03{\pm}07.63$ \\
        & RRT-MIL  & $81.93{\pm}01.71$ & $65.82{\pm}09.33$ \\
        & ILRA     & $79.64{\pm}05.73$ & $68.91{\pm}08.14$ \\
        & TransMIL & $79.61{\pm}04.88$ & $65.02{\pm}06.89$ \\
        & DFTD-MIL & $72.04{\pm}03.98$ & $64.16{\pm}05.22$ \\
        \bottomrule
    \end{tabular}%
    }
\end{table}

\noindent\textbf{TCGA results.}
Tables~\ref{tab:tcga_coad_results} and~\ref{tab:tcga_kirc_results} test generalization of the spatial features on public datasets against the full set of baselines. For COAD, DTMF-MIL-Nyström with UNI and DTMF-MIL-Local with UNI2H achieve the highest displayed mean AUC and BAcc. For KIRC, DTMF-MIL-Swin with UNI achieves the highest mean AUC and BAcc, while DTMF-MIL-Local with UNI2H achieves the highest displayed mean AUC but not BAcc. The spatial models still use $K=5$. We also report the results on coarser K=2 partition for TCGA cohorts in Supplementary Section \ref{sec:tcga_k2}.

\subsection{Spatial-Partition Analysis}
\begin{table}[t]
    \centering
    \scriptsize
    \setlength{\tabcolsep}{2.0pt}
    \renewcommand{\arraystretch}{1.05}
    \caption{\textbf{Two-region SDF analysis on IPC data with UNI2H.} We compare K-means $K=2$ with model-derived epithelial/stromal segmentation for the DTMF-MIL-Local model.}
    \label{tab:k2_boundary_results}
    \begin{tabular}{@{}lccc@{}}
        \toprule
        Partition & AUC$\uparrow$ & BAcc$\uparrow$ & $\kappa{\times}100$ \\
        \midrule
        Segmentation-2 & $\mathbf{91.33{\pm}00.54}$ & $\mathbf{82.20{\pm}01.02}$ & $\mathbf{64.18{\pm}02.06}$ \\
        K-means $K=2$ & ${91.33{\pm}00.56}$ & ${81.91{\pm}01.03}$ & ${63.64{\pm}02.19}$ \\
        \bottomrule
    \end{tabular}
\end{table}

\noindent\textbf{Two-region partitions.}
Table~\ref{tab:k2_boundary_results} shows that both K-means $K=2$ and model-derived epithelial/stromal segmentation yield competitive spatial inputs for the DTMF-MIL-Local model on the UNI2H backbone. The partitions yield identical displayed mean AUCs and similar BAcc and $\kappa$, indicating comparable performance for biologically aligned and K-means-derived regions. The two-region comparison follows the main evaluation protocol, using five random seeds and patient-level three-fold cross-validation, with results aggregated across the five seed-level cross-validation summaries.

\begin{table}[t]
    \centering
    \scriptsize
    \setlength{\tabcolsep}{2.2pt}
    \renewcommand{\arraystretch}{1.05}
    \caption{\textbf{Partition granularity ablation on IPC, UNI2H.} Values are mean $\pm$ standard deviation on DTMF-MIL-Local model.}
    \label{tab:k_sweep_results}
    \begin{tabular}{@{}lccc@{}}
        \toprule
        Partition & Dim. & AUC$\uparrow$ & BAcc$\uparrow$ \\
        \midrule
        $K=4$ & 36 & $\underline{91.90{\pm}01.33}$ & $\underline{83.48{\pm}02.19}$ \\
        $K=5$ & 45 & $\mathbf{92.18{\pm}00.19}$ & $\mathbf{85.17{\pm}01.18}$ \\
        $K=6$ & 54 & $90.85 {\pm}01.80$ & $82.02{\pm}01.18$\\
        $K=7$ & 63 & $90.94{\pm}00.83$ & $82.26{\pm}00.54$ \\
        $K=8$ & 72 & $90.61{\pm}01.00$ & $82.09{\pm}01.46$ \\
        $K=9$ & 81 & $91.01{\pm}01.04$ & $81.81{\pm}01.85$ \\
        \bottomrule
    \end{tabular}
\end{table}

\noindent\textbf{Partition granularity.}
Table~\ref{tab:k_sweep_results} shows that the useful spatial representation is consistent across varying partition granularities. On UNI2H, the default $K=5$ partition gives the best AUC and BAcc, suggesting that richer abstract tissue partitions can capture metastatic-risk structure beyond a coarse two-compartment boundary. As K-partition increases similar phenotypes are separated into different clusters leading to drop in performance. The partition-granularity and spatial-feature-control experiments use a single seeded three-fold cross-validation run, with the mean and standard deviation computed across folds.

\begin{table}[t]
    \centering
    \footnotesize
    \renewcommand{\arraystretch}{1.05}
    \caption{\textbf{Spatial-feature ablations on IPC with UNI2H.} Values are the mean $\pm$ standard deviation on DTMF-MIL-Local model.}
    \label{tab:feature_ablation_auc}
    \begin{tabular}{@{}lc@{}}
        \toprule
        Variant & AUC$\uparrow$ \\
        \midrule
        Zero spatial input & $90.46{\pm}01.40$ \\
        RBF only($r_i$) & $90.73{\pm}00.89$ \\
        RBF and SDF statistics ($r_i^k,\mu_i^k,\sigma_i^k$) & $90.95{\pm}00.89$ \\
        % RBF, SDF statistics and Gradient magnitude ($r_i^k,\mu_i^k,\sigma_i^k$,||$g_{x,i}^k$,$g_{y,i}^k$||) & $91.05{\pm}01.02$ \
        RBF, SDF statistics and gradients ($r_i^k,\mu_i^k,\sigma_i^k$,$g_{x,i}^k$,$g_{y,i}^k$) & $\mathbf{92.18{\pm}00.19}$ \\
        % \midrule
        % \midrule
        % Visual-only branch(differ) & $\underline{91.96{\pm}00.49}$ \\
        \bottomrule
    \end{tabular}
\end{table}

\noindent\textbf{Feature controls} on the UNI2H backbone are presented in Table~\ref{tab:feature_ablation_auc}. With the same PEG module, zeroing spatial inputs also reduces AUC. Removing gradients, or using only RBF responses lowers AUC relative to the full SDF descriptor. These ablations support the interpretation that the gain in performance in metastasis prediction is not simply from architectural modifications.

\section{Discussion and Conclusion}
\label{sec:discussion}

Metastatic progression is shaped not only by the appearance of individual tumor regions, but also by how tumor and surrounding tissue are arranged \cite{cancer_stroma_crosstalk_colonization,metastatic_colonization,metastasisandphenotypes}. Rather than treating a WSI as an unordered collection of patches, DTMF-MIL gives each patch a coordinate in a tissue-geometry field  through its signed distance to phenotypic regions, its direction relative to nearby interfaces, and the local statistics of the distance field. The MIL attention module then reasons over patch feature evidence after it has been contextualized by this spatial description. 

DTMF-MIL shows strong performance across both UNI and UNI2H features on three diverse datasets showcasing the utility of the phenotype derived distance features. Two-region SDFs built from K-means or epithelial/stromal segmentation perform strongly, supporting the biological intuition that coarse tissue interfaces contain metastatic-risk information. At the same time, the $K$ sweep shows that a richer $K=5$ partition can improve performance, indicating that the useful signal is not limited to a single stroma--epithelium boundary. In practice, DTMF-MIL performs well with either coarse, biologically interpretable partitions or more abstract multi-region partitions.

The feature ablations point in the same direction. RBF distance responses, directional gradients, and local SDF statistics each describe a different aspect of tissue geometry: proximity to regions, orientation around boundaries, and neighborhood variation. Under otherwise identical architecture, performance drops when these spatial cues are removed, showing improvement doesn't come merely from adding parameters or extra channels.  A stronger biological validation in the future could come from the same formulation with pathologist-defined compartments and larger external cohorts. 

In summary, DTMF-MIL shows that distance-field representations are a practical way to bring tissue architecture into weakly supervised WSI prediction. The method keeps prediction grounded in foundational pathology model based patch features, but uses spatial context to decide how those features should be aggregated. We argue that by formulating WSI analysis as a spatial encoding problem rather than a bag-of-features classification we can model metastatic behavior better, improving tasks where outcome cannot always be relied on from a single visual instance.
{
    \small
    \bibliographystyle{ieeenat_fullname}
    \bibliography{main}
}

\clearpage
\section{Supplementary Section}
\label{sec:supplement}

\subsection{Visual Phenotypic Validation of K-Means through expert and VLM}
\label{sec:vlm_validation}

\begin{figure}
    \centering
    \includegraphics[width=\linewidth]{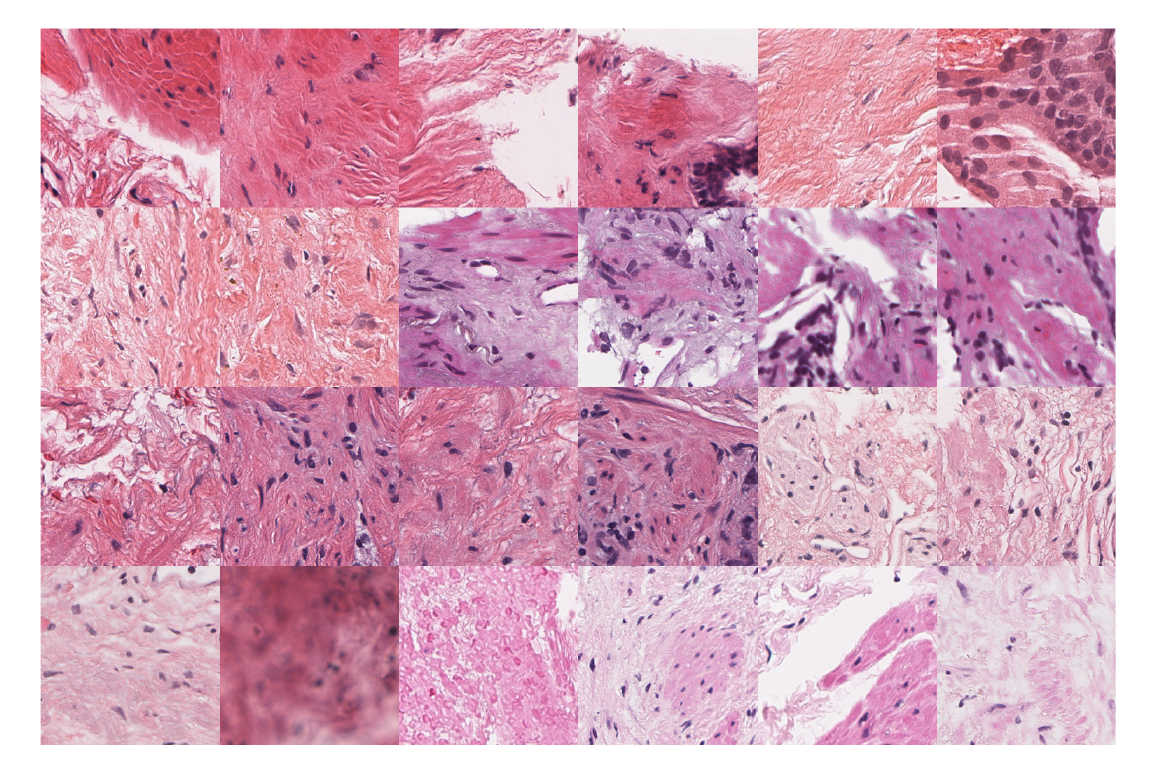}
    \caption{\textbf{Illustration of patch samples}: From cluster group 0 when $K=5$, when the VLM is prompted for the labels of these images they are interpreted as \textit{Dense Fibro-Stroma} and independently marked as \textit{Fibroblast Stroma} by a pathology expert.}
    \label{fig:cluster_sample_patches1}
\end{figure}

To evaluate whether the unsupervised data-derived tissue partitions ($K=5$) capture biologically coherent tissue structures rather than feature artifacts, we performed an independent single-patch visual morphological assessment using Gemini 1.5 Pro \cite{gemini}.

\textbf{Evaluation of $K=5$ Data-Derived Phenotypic Clusters}
We extracted 92 representative $512\times 512$ patches from the IPC dataset assigned by fold-specific K-means on UNI embeddings. The zero-shot visual evaluation confirm morphological consistency converged with these cluster identities, clusters defined by K-means and the identity by the VLM:
\begin{itemize}
    \item \textbf{Cluster 0} : Dense Fibro-Stroma 
    \item \textbf{Cluster 1} : Vascular/Inflammatory Stroma
    \item \textbf{Cluster 2} : Glandular Epithelium \& Tumor
    \item \textbf{Cluster 3} : Invasive Interface Margin
    \item \textbf{Cluster 4} : Background Void \& Artifacts
\end{itemize}

As detailed in Table~\ref{tab:vlm_cluster_phenotypes}, individual single-patch evaluation confirmed primary phenotypic alignment in \textbf{22 of 24} patches (91.7\%) for Cluster 0, \textbf{22 of 24} (91.7\%) for Cluster 1, \textbf{21 of 22} (95.5\%) for Cluster 2, and \textbf{19 of 22} (86.4\%) for Cluster 3. Overall, \textbf{84 of 92} single patch tiles (91.3\%) matched their primary cluster phenotype. 
\begin{table*}[t]
    \centering
    \scriptsize
    \setlength{\tabcolsep}{4.0pt}
    \renewcommand{\arraystretch}{1.12}
    \caption{\textbf{Single-patch visual phenotype counts for UNI foundation model K-means clusters ($K=5$) evaluated by the VLM.} Non-diagnostic background/void tiles assigned to Cluster 4 are  omitted here.}
    \label{tab:vlm_cluster_phenotypes}
    \begin{tabular}{@{}c p{3.2cm} c c p{4.2cm}@{}}
        \toprule
        \textbf{Cluster ID} & \textbf{Gemini VLM Phenotype Label} & \textbf{Matches} & \textbf{Percentage (\%)} & \textbf{Single-Patch Breakdown} \\
        \midrule
        \textbf{Cluster 0} & Dense Fibro-Stroma & 22 / 24 & 91.7\% & 22 Dense Stroma, 1 Vascular, 1 Interface \\
        \textbf{Cluster 1} & Vascular \& Inflammatory Stroma & 22 / 24 & 91.7\% & 22 Vascular, 1 Dense Stroma, 1 Interface \\
        \textbf{Cluster 2} & Glandular Epithelium / Tumor & 21 / 22 & 95.5\% & 21 Epithelium, 1 Interface \\
        \textbf{Cluster 3} & Invasive Interface Margin & 19 / 22 & 86.4\% & 19 Interface, 2 Epithelium, 1 Stroma \\
        \midrule
        \textbf{Total} & \textbf{Overall Single-Tile Alignment} & \textbf{84 / 92} & \textbf{91.3\%} & \textbf{84 exact visual matches out of 92 audited tiles} \\
        \bottomrule
    \end{tabular}
\end{table*}

\begin{figure}
    \centering
    \includegraphics[width=\linewidth]{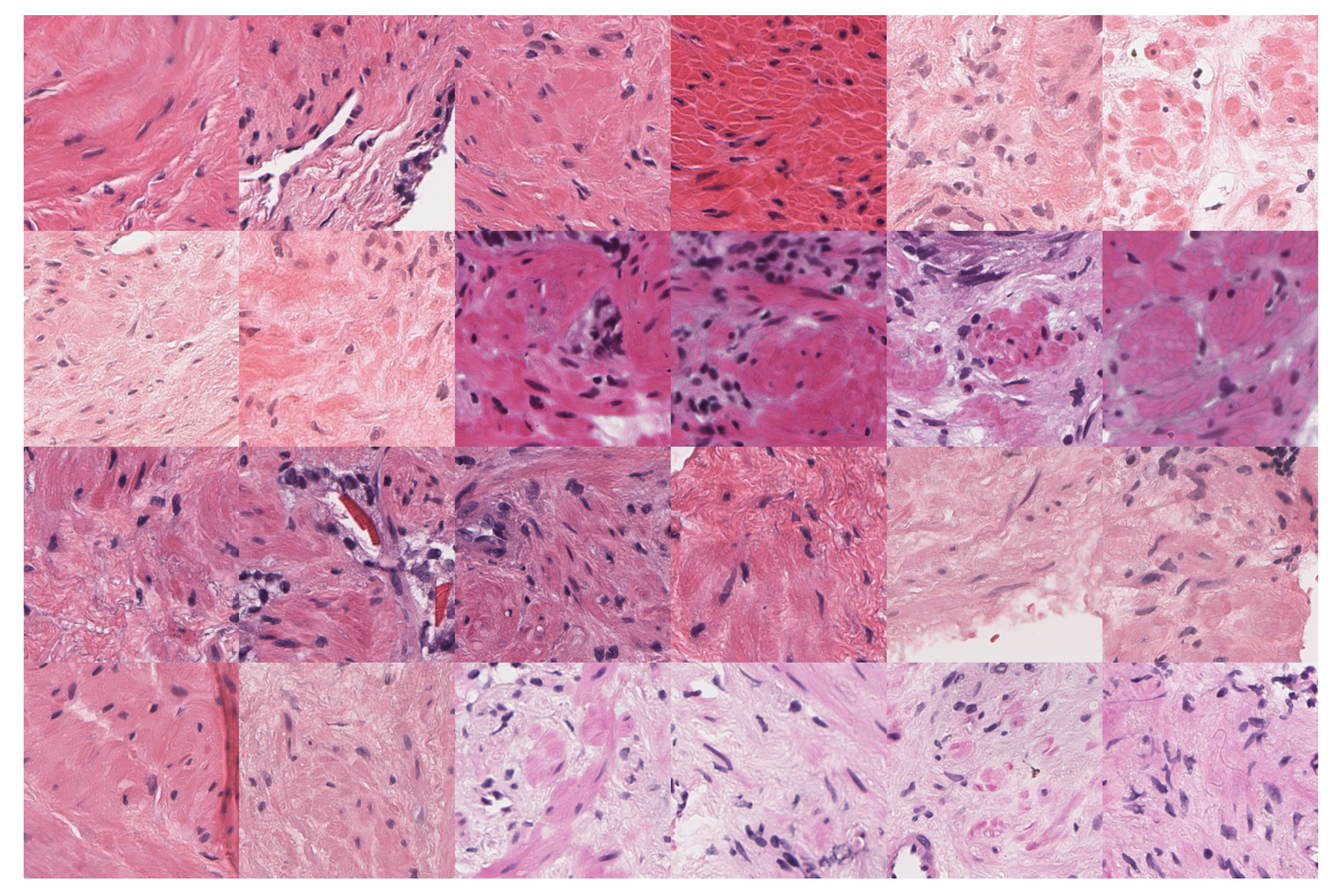}
    \caption{\textbf{Illustration of patch samples}: From cluster group 1 when $K=5$, when the VLM is prompted for the labels of these images they are interpreted as \textit{Vascular/Inflammatory Stroma} and independently marked as \textit{Fibroblast stroma and vascular structures}  by a pathology expert.}
    \label{fig:cluster_sample_patches2}
\end{figure}

\begin{figure}
    \centering
    \includegraphics[width=\linewidth]{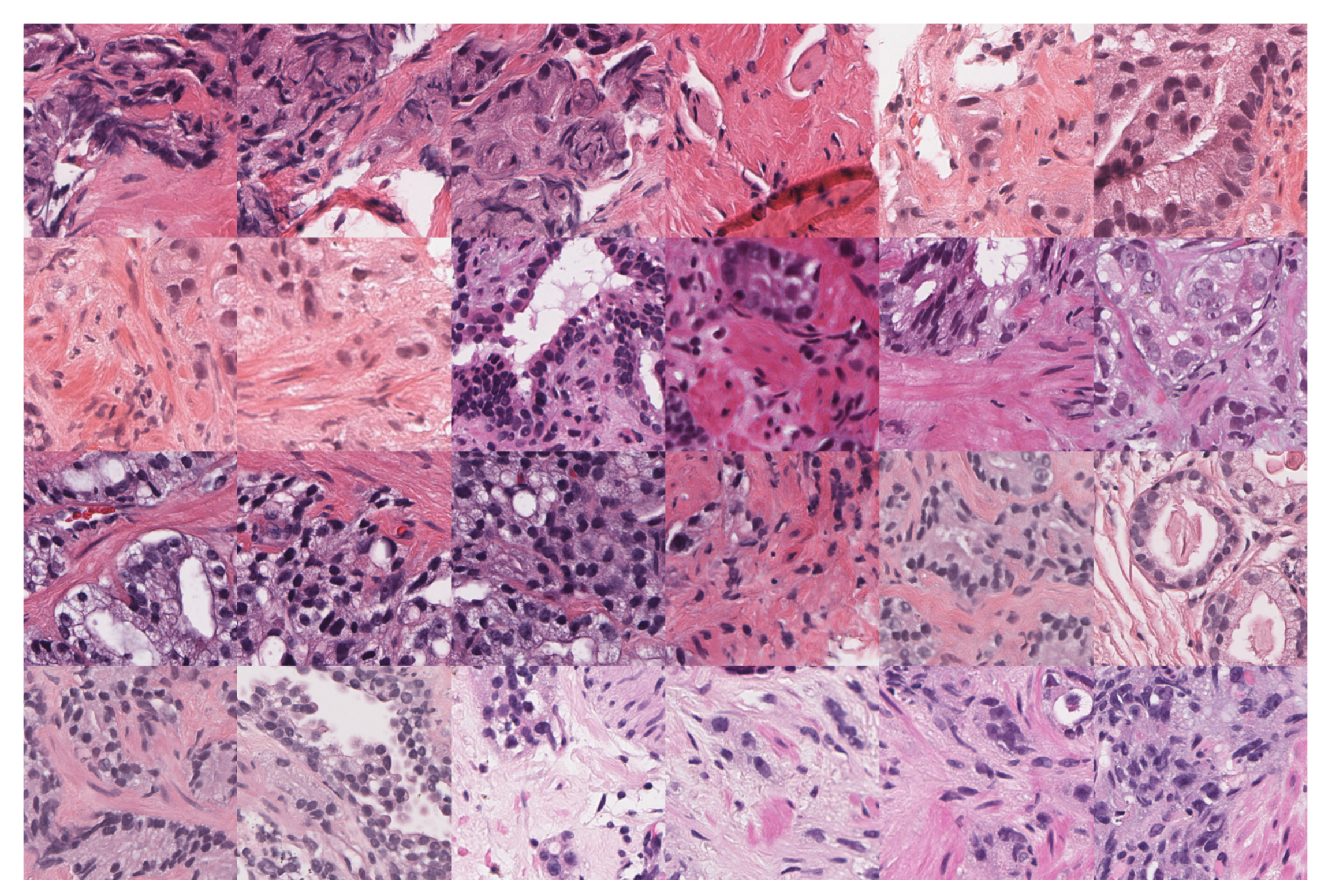}
    \caption{\textbf{Illustration of patch samples}: From cluster group 2 when $K=5$, when the VLM is prompted for the labels of these images they are interpreted as \textit{Glandular Epithelium \& Tumor} and independently marked as \textit{Benign and cancerous epithelial cells with gland formation} by a pathology expert.}
    \label{fig:cluster_sample_patches3}
\end{figure}

\begin{figure}
    \centering
    \includegraphics[width=\linewidth]{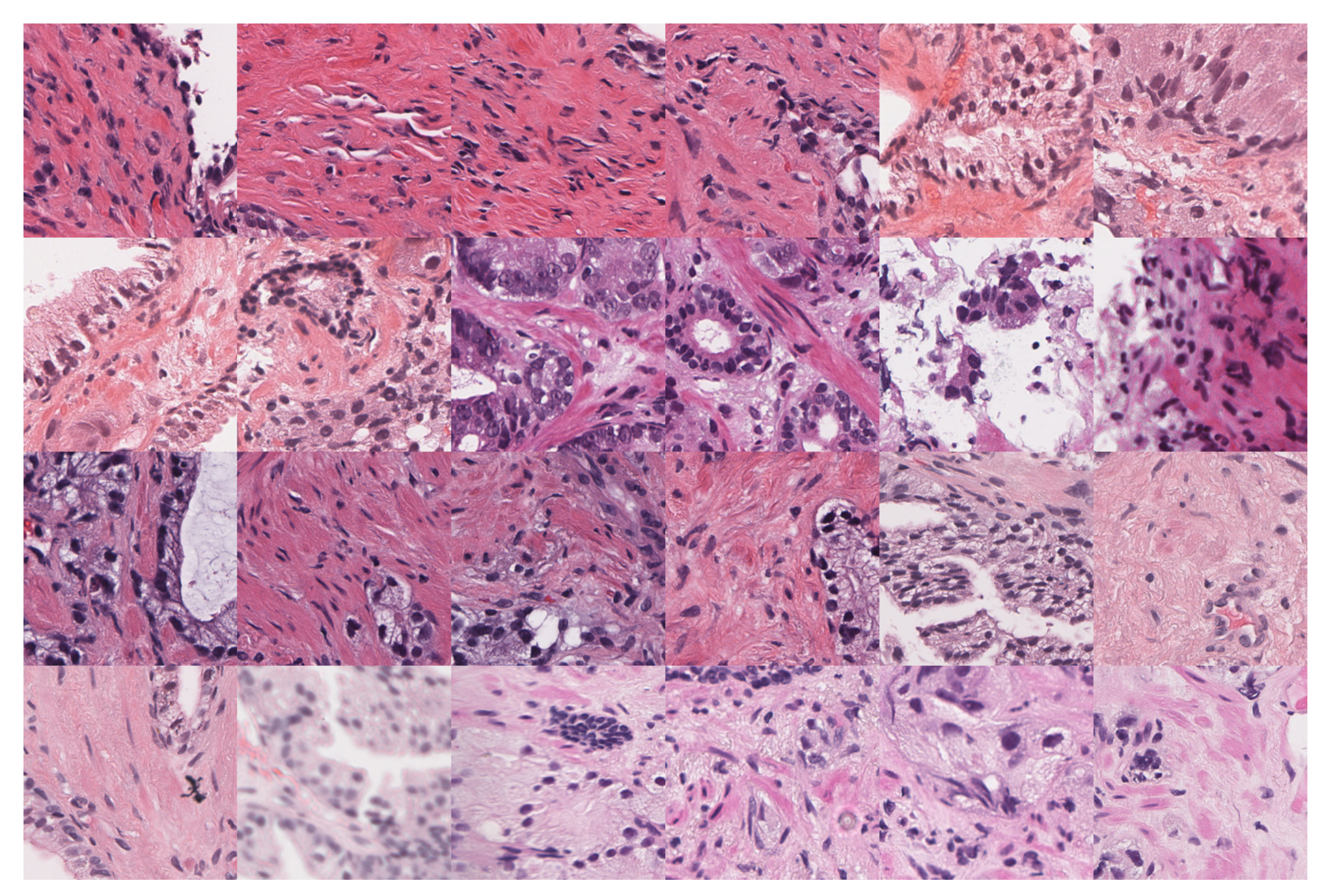}
    \caption{\textbf{Illustration of patch samples}: From cluster group 3 when $K=5$, when the VLM is prompted for the labels of these images they are interpreted as \textit{Invasive Interface Margin \& Transition Patches} and independently marked as \textit{High Grade Cancer Nests} by a pathology expert.}
    \label{fig:cluster_sample_patches4}
\end{figure}

Further a certified pathologist reviewed the cluster images shown in figures. \ref{fig:cluster_sample_patches1}, \ref{fig:cluster_sample_patches2}, \ref{fig:cluster_sample_patches3}, \ref{fig:cluster_sample_patches4} and identified the groups as:
\begin{itemize}
    \item \textbf{Cluster 0} : Fibroblast Stroma
    \item \textbf{Cluster 1} : Fibroblast stroma and vascular structures
    \item \textbf{Cluster 2} : Benign and cancerous epithelial cells with gland formation
    \item \textbf{Cluster 3} : High grade cancer nests
\end{itemize}
showing an alignment with the VLM judgement and the K-means tissue partitions.

\subsection{Segmentation Model}
\label{sec:segmentation_model}
Glandular structures were delineated with a segmentation model provided by Cai. et al. ~\cite{segmentationmodel}, using their released code base: a DeepLabV3+ network~\cite{chen2018deeplabv3plus} with a ResNet-101 backbone, trained on the public RINGS prostate-biopsy gland segmentation dataset~\cite{salvi2021rings}. Following the original protocol the gland-boundary class is merged into stroma after the arg-max, giving a binary gland vs.\ non-gland segmenter, which reaches a pixel accuracy of $0.963$ and an mIoU of $0.926$ on the held-out RINGS test split. We applied this model to
our cohort without fine-tuning: individual biopsy cores were localized on each slide by color-based tissue detection and connected-component analysis, and
each core was segmented at $20\times$ ($0.488\,\mu$m/px), tiled where necessary to bound GPU memory. Each core was segmented as a complete rectangular crop, since blanking the glass beforehand degraded the predictions; non-tissue pixels were discarded afterwards by intersecting the predicted label maps with a tissue mask obtained from the same color rules, so all reported gland areas and fractions are measured within tissue only.

\subsection{Scaling Mean SDFs}
\label{sec:scaling}
Because the mean SDF values we compute are in the range of $-10^3$ to $10^3$ due to the large sizes of WSIs, we scale these values so that the network is stable during training. For each tissue partition $k$ and patch $p_i$, the raw descriptor contains RBF responses $r_i^k$, the local SDF gradients $(g_{x,i}^{k},g_{y,i}^{k})$, the local signed-distance mean $\mu_i^{k}$, and the local signed-distance standard deviation $\sigma_i^{k}$ we scale, gradients and local spread by a fixed constant depending on the size of the patch($ps$), while the local mean SDF is transformed asymmetrically:
\begin{equation}
    \tilde{g}_{x,i}^{k}=\frac{g_{x,i}^{k}}{ps}, \quad
    \tilde{g}_{y,i}^{k}=\frac{g_{y,i}^{k}}{ps}, \quad
    \tilde{\sigma}_{i}^{k}=\frac{\sigma_i^{k}}{ps}.
\end{equation}
\begin{equation}
    \tilde{\mu}_{i}^{k}=\begin{cases}
        \frac{10}{1+\mu_i^{k}/2000}, & \mu_i^{k}\ge 0,\\[3pt]
        \frac{10}{1+|\mu_i^{k}|/200}, & \mu_i^{k}<0.
    \end{cases}
\end{equation}

The asymmetric scaling keeps near-boundary regions numerically prominent while compressing large signed distances. This asymmetric transform maps both branches to $(0,10]$, while using different decay scales inside and outside region k. The shorter scale on the negative side gives stronger emphasis to one side of the local transition band, while the longer positive scale avoids collapsing broader tissue neighborhoods too aggressively. 
The resulting per-cluster descriptor is thus this scaled version,
\begin{equation}
    s_i^k = [r_{i,1}^{k},\ldots,r_{i,5}^{k},\tilde{g}_{x,i}^{k},\tilde{g}_{y,i}^{k},\tilde{\mu}_{i}^{k},\tilde{\sigma}_{i}^{k}],
\end{equation}
and the full spatial input is the concatenation of $s_i^k$ over all $K$ partitions. For all our experiments the multi scale hyperparameter $\Lambda$, where $|\Lambda| = 5$ is selected as:
\begin{equation}
\begin{aligned}
    \Lambda=\{&10^{-7},\;3.59\!\times\!10^{-7},\;1.29\!\times\!10^{-6}, \\
              &4.64\!\times\!10^{-6},\;1.67\!\times\!10^{-5}\}.
\end{aligned}
\label{eq:rbf-lambdas}
\end{equation}

\begin{table*}[t]
    \centering
    \footnotesize
    \setlength{\tabcolsep}{4pt}
    \renewcommand{\arraystretch}{1.08}
    \caption{\textbf{IPC metastatic-risk prediction with PHIKON-V2 backbone.}
    Bold indicates the best result in each metric and underline indicates the second-best result.}
    \label{tab:va_phikonv2_results}
    \begin{tabular}{lccccc}
        \toprule
        Method
        & BAcc$\uparrow$
        & AUC$\uparrow$
        & F1$\uparrow$
        & Accuracy$\uparrow$
        & $\kappa{\times}100$ \\
        \midrule

        \multicolumn{6}{l}{\textit{Spatial methods (Ours)}} \\

        \rowcolor{ourgray}
        DTMF-MIL-Local
        & $80.81 \pm 02.03$
        & $90.38 \pm 01.50$
        & $80.23 \pm 03.28$
        & $80.58 \pm 01.89$
        & $61.22 \pm 03.81$ \\

        \rowcolor{ourgray}
        DTMF-MIL-Swin
        & $81.13 \pm 02.15$
        & $90.01 \pm 01.54$
        & $\underline{81.01 \pm 02.78}$
        & $\underline{80.93 \pm 02.01}$
        & $\underline{61.90 \pm 04.03}$ \\

        \rowcolor{ourgray}
        DTMF-MIL-Nystrom
        & $\mathbf{81.80 \pm 02.13}$
        & $\mathbf{90.62 \pm 00.97}$
        & $\mathbf{82.15 \pm 02.31}$
        & $\mathbf{81.66 \pm 02.05}$
        & $\mathbf{63.31 \pm 04.08}$ \\

        \multicolumn{6}{l}{\textit{Non-spatial baselines}} \\
        \hline

        CLAM
        & $\underline{81.17 \pm 02.97}$
        & $\underline{90.58 \pm 00.71}$
        & $80.37 \pm 04.65$
        & $80.72 \pm 03.18$
        & $61.71 \pm 06.09$ \\

        WiKG
        & $78.22 \pm 06.24$
        & $90.05 \pm 01.62$
        & $74.59 \pm 12.66$
        & $77.73 \pm 06.75$
        & $55.92 \pm 12.71$ \\

        ABMIL
        & $79.62 \pm 03.75$
        & $89.88 \pm 01.55$
        & $79.66 \pm 05.03$
        & $79.21 \pm 04.05$
        & $58.70 \pm 07.76$ \\

        DSMIL
        & $80.23 \pm 03.19$
        & $90.08 \pm 01.59$
        & $80.17 \pm 04.25$
        & $80.03 \pm 03.47$
        & $60.15 \pm 06.60$ \\

        RRT-MIL
        & $80.18 \pm 04.67$
        & $90.05 \pm 01.34$
        & $78.28 \pm 08.01$
        & $79.64 \pm 05.29$
        & $59.73 \pm 09.74$ \\

        ILRA
        & $78.40 \pm 04.80$
        & $88.13 \pm 02.81$
        & $76.10 \pm 09.49$
        & $78.27 \pm 05.03$
        & $56.60 \pm 09.64$ \\

        TransMIL
        & $78.38 \pm 04.07$
        & $87.72 \pm 01.84$
        & $76.96 \pm 07.50$
        & $78.12 \pm 04.44$
        & $56.42 \pm 08.37$ \\

        DFTD-MIL
        & $73.20 \pm 09.03$
        & $79.64 \pm 09.66$
        & $67.24 \pm 21.37$
        & $72.79 \pm 09.56$
        & $46.10 \pm 18.01$ \\

        \bottomrule
    \end{tabular}
\end{table*}

\subsection{Implementation Details}
\label{sec:implementation_details}
Each WSI is represented as a bag of patch embeddings with associated grid coordinates. The patch size of 512x512 is used for the IPC dataset and 256x256 size patches are used for the TCGA dataset both at 20x magnification. For the main method, K-means provides data-driven tissue partitions from which per-region SDFs are computed. K-means is fitted separately within each cross-validation fold using only training-patient slides, and the resulting fold-specific centroids are applied to validation and held-out patients. We further compare K-means $K=2$ with epithelial/stromal partitions estimated by the unsupervised histopathology semantic segmentation model of Cai et al.~\cite{segmentationmodel} trained on a prostate dataset. This comparison asks whether a coarse, biologically recognizable tissue interface is itself informative. The models were trained on NVIDIA H200 GPU. We trained with binary cross entropy loss, AdamW optimization setting at learning rate $10^{-3}$ and weight decay at $10^{-4}$. The models were trained for 20 epochs with best model selected from the validation result. Results are averaged over the same three folds and five random seeds used in the main experiments. The code and features will be made publicly available for reproducibility.

\subsection{IPC Phikon-V2}
\label{sec:ipc_phikon-v2}
Results on foundation model Phikon-V2 based backbone for the IPC dataset are presented in the table \ref{tab:va_phikonv2_results}. 
When comparing DTMF-MIL-Local for each backbone, with other five representative baselines (CLAM, WiKG, ABMIL, DSMIL, and RRT-MIL) across five metrics (AUC, F1 score, accuracy, balanced accuracy, and $\kappa$), yielding \(5\text{ baselines}\times5\text{ metrics}=25\) pairwise comparisons per backbone, DTMF-MIL-Local achieved a higher mean value in 20 of 25 comparisons with Phikon-v2, 24 of 25 comparisons with UNI, and all 25 comparisons with UNI2H. Across the three backbones, this yielded \(3\text{ backbones}\times25\text{ comparisons}=75\) comparisons in total, of which 69 favored DTMF-MIL-Local.

\subsection{TCGA K=2 Ablations}
\label{sec:tcga_k2}
The main TCGA experiments use $K=5$ tissue partitions. To test whether a coarser two-region partition can also carry useful spatial signal beyond the IPC prostate setting, we run DTMF-MIL with K-means $K=2$ on TCGA-COAD and TCGA-KIRC. The results are presented in Table. \ref{tab:supp_tcga_k2_coad} and \ref{tab:supp_tcga_kirc_k2}. Values are multiplied by 100 and reported as mean $\pm$ standard deviation across the same multi-seed cross-validation protocol used in the main TCGA experiments. 

\begin{table}
    \centering
    \scriptsize
    \setlength{\tabcolsep}{2.2pt}
    \renewcommand{\arraystretch}{1.04}
    \caption{\textbf{TCGA-COAD with UNI and UNI2H.} $K=2$ spatial controls.}
    \label{tab:supp_tcga_k2_coad}
    \begin{tabular}{@{}clcc@{}}
        \toprule
        & Method & AUC$\uparrow$ & BAcc$\uparrow$ \\
        \midrule
        
        & DTMF-MIL-Local & $\mathbf{74.40{\pm}04.17}$ & $\mathbf{67.09{\pm}04.46}$ \\
        & DTMF-MIL-Swin  & $72.00{\pm}04.63$ & $65.47{\pm}06.29$ \\
        \multirow{-3}{*}{\rotatebox[origin=c]{90}{UNI}} & DTMF-MIL-Nys.  & $73.88{\pm}04.42$ & $66.14{\pm}05.68$ \\
        
        \midrule \midrule
        & DTMF-MIL-Local & $72.15{\pm}04.08$ & $63.91{\pm}05.05$ \\
        & DTMF-MIL-Swin  & $\mathbf{72.74{\pm}04.51}$ & $\mathbf{65.61{\pm}05.93}$ \\
        \multirow{-3}{*}{\rotatebox[origin=c]{90}{UNI2H}} & DTMF-MIL-Nys. & $71.07{\pm}05.61$ & $65.47{\pm}04.86$ \\
        
        \bottomrule
    \end{tabular}
\end{table}

\begin{table}
    \centering
    \scriptsize
    \setlength{\tabcolsep}{2.2pt}
    \renewcommand{\arraystretch}{1.04}
    \caption{\textbf{TCGA-KIRC with UNI and UNI2H} $K=2$ spatial controls.}
    \label{tab:supp_tcga_kirc_k2}
    \begin{tabular}{@{}clcc@{}}
        \toprule
        & Method & AUC$\uparrow$ & BAcc$\uparrow$ \\
        \midrule
        & DTMF-MIL-Local & $83.67{\pm}02.06$ & $70.00{\pm}06.72$ \\
        & DTMF-MIL-Swin & $\mathbf{84.13{\pm}02.48}$ & $\mathbf{71.19{\pm}07.92}$ \\
        \multirow{-3}{*}{\rotatebox[origin=c]{90}{UNI}} & DTMF-MIL-Nys. & $82.15{\pm}02.91$ & $69.75{\pm}05.55$ \\
        
        \midrule \midrule
        & DTMF-MIL-Local & $82.16{\pm}03.10$ & $67.78{\pm}07.79$ \\
        & DTMF-MIL-Swin & $\mathbf{82.28{\pm}03.14}$ & $\mathbf{70.84{\pm}05.76}$ \\
        \multirow{-3}{*}{\rotatebox[origin=c]{90}{UNI2H}} & DTMF-MIL-Nys. & $82.27{\pm}02.14$ & $67.87{\pm}07.80$ \\
        
        \bottomrule
    \end{tabular}
\end{table}

Across TCGA, the $K=2$ setting remains competitive but does not uniformly dominate the default $K=5$ partition. In COAD with UNI, the DTMF-MIL-Local model improves over its $K=5$ counterpart in both AUC and BAcc, while COAD with UNI2H is weaker for the local model and stronger for the DTMF-MIL-Swin. In KIRC, the $K=2$ DTMF-MIL-Swin is particularly strong, especially on BAcc. Overall, the result supports the idea that coarse tissue-interface geometry can be informative, while the default $K=5$ representation gives a more flexible partition for the primary method.

\subsection{Attention Visualization}
\begin{figure*}
    \centering
    \includegraphics[width=\linewidth]{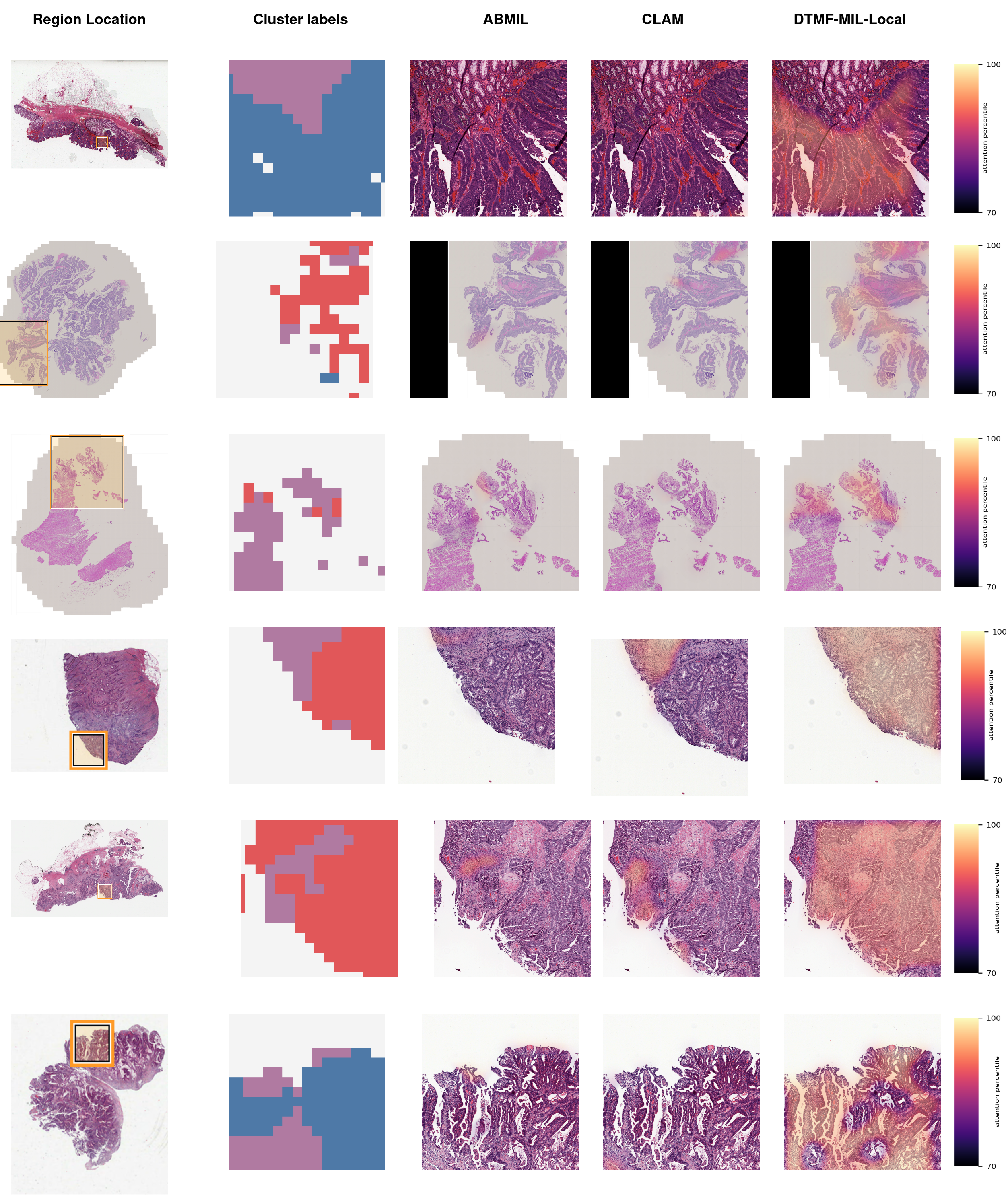}
    \caption{Qualitative attention comparison on six WSIs (TCGA-COAD). Each row shows the source-region location, K-means patch assignments, and attention from ABMIL, CLAM, and DTMF-MIL-Local on the same crop.}
    \label{fig:attn_viz}
\end{figure*}

\begin{figure*}
    \centering
    \includegraphics[width=\linewidth]{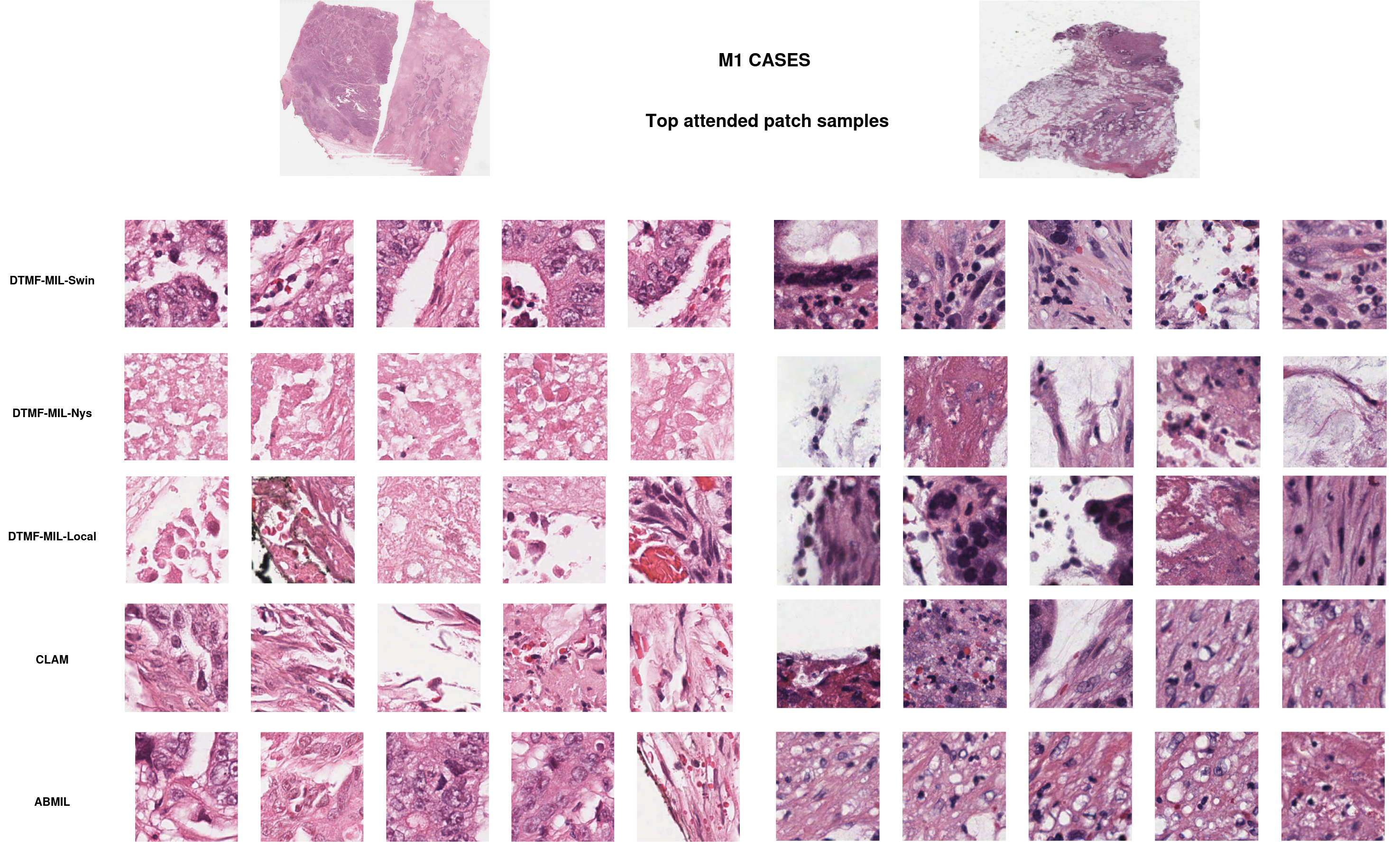}
    \caption{Highest-attention patches from M1-labeled WSIs under DTMF-MIL-Local. Within each row, patches are ordered from highest to lowest attention. Patch size is 256x256, and the overview identifies the source WSI.}
    \label{fig:sup_viz1}
\end{figure*}

\begin{figure*}
    \centering
    \includegraphics[width=\linewidth]{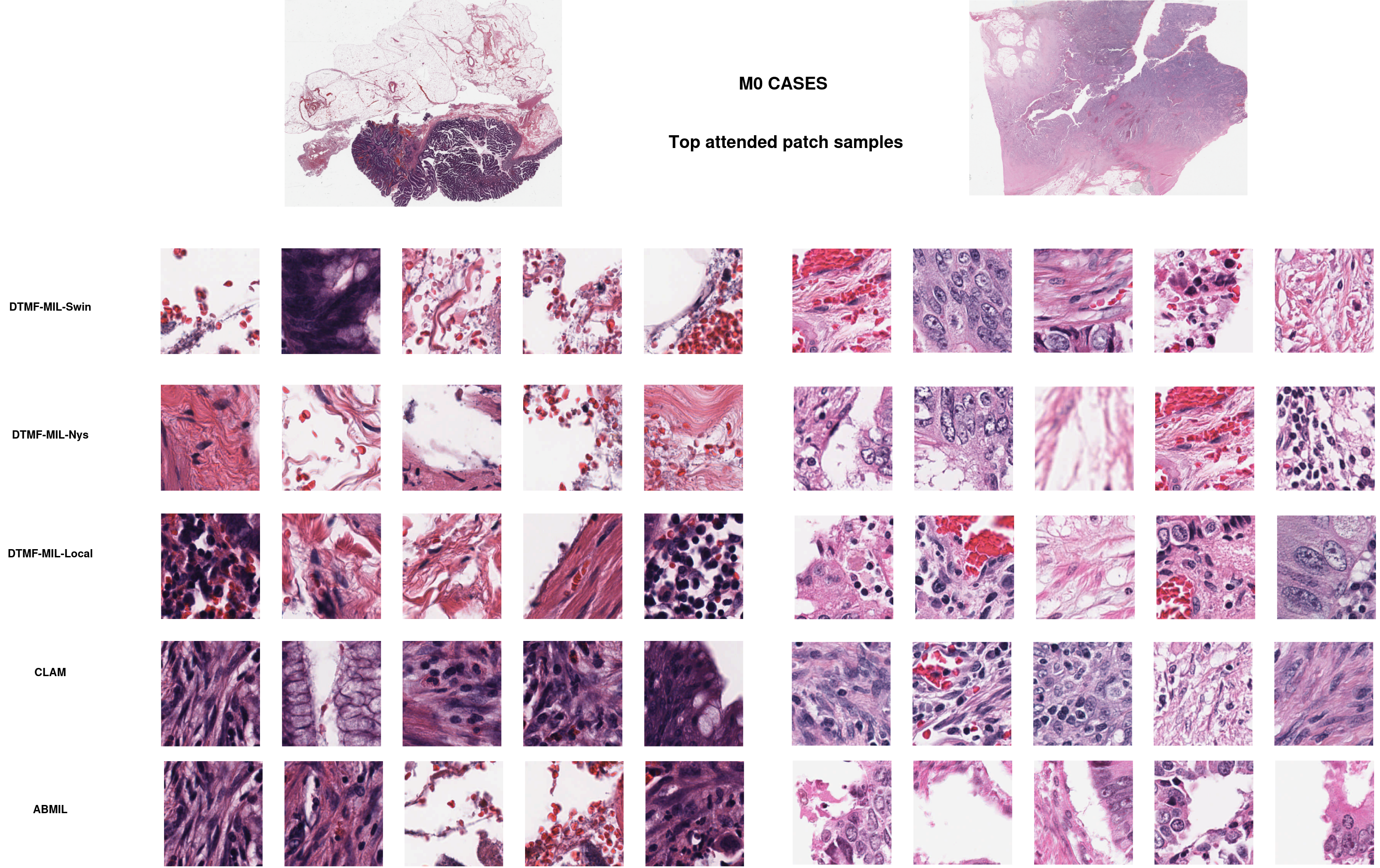}
    \caption{Highest-attention patches from M0-labeled WSIs under DTMF-MIL-Local. Within each row, patches are ordered from highest to lowest attention. Patch size is 256x256, and the overview identifies the source WSI.}
    \label{fig:sup_viz2}
\end{figure*}

Figure. \ref{fig:attn_viz}, illustrates the Whole Slide Image and a cropped region of the WSI, the patch cluster labels assigned by the K-means model, and the attention in those regions by AMBIL, CLAM and DTMF-MIL-Local models. The cropped regions are made around adjacent cluster regions and the presented samples show increased attention of DTMF-MIL model around these regions compared to Conventional MIL models. Additionally figures \ref{fig:sup_viz1} and \ref{fig:sup_viz2} show the highest-attention patches from M1-labeled and M0-labeled WSIs, respectively. 

\label{sec:attn_viz}

\subsection{Limitations and Future Work}
\label{sec:limitations} 

DTMF-MIL currently uses data-derived tissue partitions without requiring pathologist-provided region annotations, which supports scalability but limits the direct semantic interpretation of the clusters. Future work could use vision-language models to characterize or assign histologic phenotypes to the learned clusters, followed by pathologist validation, although this would introduce additional computational cost and potential annotation noise. The current pipeline is also not fully end-to-end because SDF construction, and MIL optimization are performed as separate stages. A differentiable clustering mechanism could allow tissue partitions and spatial descriptors to be learned jointly for the prediction task. Similarly, the scales used by the radial-basis distance responses are currently fixed hyperparameters, learning these scales could allow the model to adapt its spatial sensitivity to different tissue structures and cancer types.

Evaluation on larger, multi-institutional cohorts is also necessary to establish robustness across patient populations, specimen-preparation protocols, and scanning conditions. However, large, well-curated public datasets that pair primary-tumor WSIs with reliable patient-level distant-metastasis labels remain limited. Although private institutional cohorts can support studies in this area, restricted access prevents independent reproduction and standardized comparison across methods. TCGA is one of the few publicly accessible resources that permits such analysis by linking pathology slides with TNM metadata, although M-stage availability and label quality vary across cases and cancer types. Additional publicly accessible, multi-institutional benchmarks would therefore support more reliable and reproducible evaluation and could encourage broader investigation of this task. Importantly, the present study predicts existing M0 versus M1 status from primary-tumor tissue and does not estimate whether an M0 tumor will develop metastasis in the future. The intended direction is to investigate whether primary-tissue architecture can provide complementary evidence of distant disease before metastatic-site tissue is available.

\end{document}